# Ma(r)king concessions in English and German

Brigitte Grote
FAW Ulm
Helmholtzstr. 16
D-89081 Ulm
grote@faw.uni-ulm.de

Nils Lenke
Universität Duisburg
FB 3, FG Computerlinguistik
D-47048 Duisburg
he233le@unidui.uni-duisburg.de

Manfred Stede
University of Toronto and
FAW Ulm
Helmholtzstr. 16
D-89081 Ulm
stede@faw.uni-ulm.de

**Abstract**

In order to generate cohesive discourse, many of the relations holding between text segments need to be signalled to the reader by means of cue words, or *discourse markers*. Programs usually do this in a simplistic way, e.g., by using one marker per relation. In reality, however, language offers a very wide range of markers from which informed choices should be made. In order to account for the variety and to identify the parameters governing the choices, detailed linguistic analyses are necessary. We worked with one area of discourse relations, the CONCESSION family, identified its underlying pragmatics and semantics, and undertook extensive corpus studies to examine the range of markers used in both English and German. On the basis of an initial classification of these markers, we propose a generation model for producing bilingual text that can incorporate marker choice into its overall decision framework.

# 1 Introduction

The existence of a CONCESSION[1] category is acknowledged by standard grammars as well as in work on rhetorical relations (e.g., Mann and Thompson 1987). As for defining it, there is general agreement that some kind of "failed expectation" is involved; for example, Quirk et al. (1972) say that "concessive conjuncts signal the unexpected, surprising nature of what is being said in view of what was said before that." Similarly, Helbig and Buscha (1991) characterize a concessive sentence as one where an expected causal relationship does not hold—a cause given in a subordinate clause does not have the consequence one would anticipate from a law of cause and effect. For Halliday (1985), CONCESSION is a subtype of CONDITION, its meaning being "if P then contrary to expectation Q". Martin (1992) sees CONCESSION as a cross-classification of various CAUSE relations, and others see it as a subtype of CONTRAST or ADVERSATIVE (Halliday and Hasan 1976, Lang 1989). On the other hand, Mann and Thompson (1987) define CONCESSION on a rhetorical level, having to do with increasing the reader's positive regard for a proposition (see the full definition in figure 1).

Given this variety of definitions, it is not surprising to encounter a wide range of example cases in the literature, which are supposed to illustrate CONCESSION, and with it a wealth of linguistic forms signalling concession:

(1) *Selbst wenn er sich noch so sehr anstrengt, wird er dennoch nicht Präsident werden.* (Bußmann 1990)

(2) *Er arbeitet, obwohl er schon alt ist.* (Helbig and Buscha 1991)

(3) *Es regnete zwar in Strömen, aber wir gingen trotzdem spazieren.* (Helbig and Buscha 1991)

(4) *Trotz des schlechten Wetters gingen wir spazieren.* (Helbig and Buscha 1991)

(5) *Despite strong pressure from the government, the unions have refused to order return to work.* (Quirk et al. 1972)

(6) *Their term papers were very brief. Still, they were better than I expected.* (Quirk et al. 1972)

(7) *Although the material is toxic to certain animals, evidence is lacking that it has any serious long-term effect on human beings.* (Mann and Thompson 1987)

(8) *Possibly, but indeed I don't know, although they stood whispering very near to me: because they stood at the top of the cabin steps to have the light of the lamp that was hanging there; it was a dull lamp, and they spoke very low, and I did not hear what they said, and saw only that they looked at papers.* (C. Dickens: A Tale of Two Cities)

---

[1]Typeface convention: Throughout the paper, we reserve SMALLCAPS for names of discourse relations, *slant* for linguistic examples, and *italics* for emphasis.

(9)   *Er hat zwar kein Auto, aber dafür ein Fahrrad.*

(10)

(a)   *Und ob ich schon wanderte im finstern Tal, fürchte ich kein Unglück; denn du bist bei mir, dein Stecken und Stab trösten mich.* (Psalm 23, Luther-Bibel)

(b)   *Yea, though I walk through the valley of the shadow of death, I will fear no evil: for thou art with me; thy rod and thy staff they comfort me.* (Psalm 23, King James Bible)

In short, there is no clear and agreed-upon definition of a CONCESSION relationship, and accordingly it is hard to motivate classifications of the linguistic *cues* that signal the presence of such a relation and allow an informed choice among a set of candidate markers and constructions.

Essentially, two things are missing. On the one hand, a broader definition covering the various cases of concessions; on the other hand, something like the equivalent of Roget's thesaurus for function words: the mapping from meaning to linguistic expression, providing us with factors determining marker choice.

¿From the perspective of language generation, one wants to verbalize a "deep" representation of content and produce coherent as well as cohesive text. Hence, both tasks just outlined need to be resolved, if automatic language generators are expected to move beyond trivial rules like always mapping the "concession" feature to the subordinators *although* and *obwohl*, respectively (as, for example, the PENMAN generator (Penman 1989) does). This involves specifying the different levels of representation (intentions, discourse structure, semantics) and identifying the meanings of the markers. In NL generation, this area has not seen many results yet. Exceptions are Elhadad and McKeown (1990), who investigate specific cases like the difference between producing clauses with *although* and *but*, and Scott and de Souza (1990), who propose a set of general heuristics for mapping a discourse representation tree to language.

Furthermore, we approach the task from a bilingual perspective, trying to find the common ground of both English and German realizations of CONCESSION and to compare them. Looking at more than one language broadens the perspective on the phenomenon and, from theoretical as well as practical viewpoints, leads to language-neutral levels of representation.

In the following, our overall goal is an account of the correlation between linguistic realizations of CONCESSION and those features of discourse situations that influence the choice among them – roughly speaking, a "grammar of conceding". Our approach can be summarized as follows:

- We propose three different *pragmatic* classes of making concessions in discourse, corresponding to communicative goals (section 2).

- We identify a single *semantic* representation scheme underlying all concessions, in terms of believed propositions and default rules (section 3).

1. **relation name:** CONCESSION
2. **constraints on N:** W has positive regard for the situation presented in N;
3. **constraints on S:** W is not claiming that the situation presented in S doesn't hold;
4. **constraints on the N + S combination:** W acknowledges a potential or apparent incompatibility between the situations presented in N and S; W regards the situations presented in N and S as compatible; recognising the compatibility between the situations in N and S increases R's positive regard for the situation presented in N;
5. **the effect:** R's positive regard for the situation presented in N is increased
6. **locus of the effect:** N and S

   (N = nucleus, S = satellite, W = writer, R = reader)

Figure 1: Definition of CONCESSION in (Mann and Thompson 1987)

- We collect a variety of *syntactic* and *lexical* realizations of CONCESSION in both English and German, and make observations on typical orderings of concessive statements in discourse (section 4).

- We propose a model of (bilingual) generation that maps a configuration of communicative goals, beliefs, and presuppositions first to an RST-like discourse representation, and then to language-specific semantic sentence representations. These are given to surface generation modules, which produce linguistic output (section 5).

## 2 Why do we concede?

Looking for the different reasons for making concessions in discourse, we followed three paths. First, the dictionary definitions of concessive markers found in Cobuild (1987) and Duden (1989) provided a few distinctions; similarly, we inspected classifications of discourse relations in grammars (Quirk et al. 1972; Helbig and Buscha 1991) and the literature (e.g., Martin 1992, Mann and Thompson 1987) and compared them to the other findings. Finally, for every marker we gathered concordances from online corpora (LIMAS (Glas 1975); COBUILD) and categorized the examples into groups reflecting the reasons for making the statement. Contrary to the treatment of CONCESSION in RST, which emphasizes the goal of increasing the reader's positive regard for a situation, we found more rationales for conceding. Thus, the following three major groups emerged, which are distinct with respect to the motivation for conceding.[2]

---

[2]The borders between these groups are by no means clear-cut, though, since all concessive statements have something in common (recall the element of "failed expectation" mentioned in the

## 2.1 Concede-I: convince the hearer

When the goal is to convince the hearer of a particular point, or get him or her to act in a particular way, a common rhetorical strategy is to paraphrase counter-arguments already mentioned, or to anticipate those not yet uttered, and to concede them, while at the same time insisting on the dominance of one's own argument.

(11) *Although you are correct that Windows is cheap I nevertheless wouldn't buy it, because it has many bugs.*

This "argumentative" type of CONCESSION corresponds most closely to the definition by Mann and Thompson (1987) given in figure 1.

As mentioned in some of the definitions in the literature, CONCESSION is a specific case of stating a CONTRAST, where two propositions are presented whose co-occurrence is for some reason unusual:

(12) *Charly is quite tall, while his brother is short.*

General contrastive markers like *however*, *but* or *while* can be used here. To see that making a CONTRAST can blend easily into a CONCESSION, consider:

(13) *She resembled her mother physically, though not mentally.* (Cobuild 1987)

(14) *On the one hand it is correct that Windows is cheap. But on the other hand, it has many bugs.*

*Though* can be meant concedingly, and here in some sense it is. In addition to the contrast, the two propositions are evaluated somehow, with one of them winning. Markers like *on the one hand – on the other hand*, as in the example given above, seem to imply an evaluation solely on the basis of the linear order of the arguments: there is a tendency to regard the last-mentioned clause (here *Windows has many bugs*) as the more important one. We label this distinction as the existence of a *main act* and a *minor act*[3] which is evident in (14), but appears to be absent in (12) and possibly (13) — there, none of the two propositions is more prominent than the other, since the point of the utterance is the existence of the contrast of the two.

## 2.2 Concede-II: prevent false implicatures

In order to be *cooperative* in the sense of Grice's maxims, one has to anticipate the inferences that the hearer might draw from a fact that is just introduced to the discourse. These are inferences that will be drawn using general world knowledge. However, in specific instances, the speaker wants to prevent the application of such general rules, hence the the proposition not implied is denied. As oppposed to Concede-I, the conceded fact is new to the discourse and the hearer is not assumed to hold a specific attitude towards that fact:

---

beginning).

[3]The distinction is also made by Elhadad and McKeown (1990) who note that different markers can assign different positions to main and minor act. Their example: *He failed the exam, but he is smart. Let's hire him.* — * *He failed the exam, although he is smart. Let's hire him.*.

(15) *Windows is very cheap. That doesn't mean you should buy it, though, because it is full of bugs.*

(16) *The classrooms are small, though not unsuitable.* (Cobuild 1987)

The denial of a false implicature has the character of an afterthought, which is added after the main act has been stated and the possibility of the unintended implicature comes to mind. Therefore, this kind of concession will mostly be found in spoken discourse.

## 2.3 Concede-III: inform about surprising events

A statement becomes more relevant for being communicated if it contains an unusual or surprising element that violates general expectations, or cancels an assumed cause-effect relation in "the world". This amounts to a subject-matter relation in Mann and Thompson's terms and therefore has nothing to do with increasing the hearer's positive regard or preventing false implicatures. The fact that the temperature is 20 degrees and no snow is falling is quite irrelevant (though true!) when travelling to the Mediterranean Sea in August; it is worth being reported when travelling to the Black Forest in December, though:

(17) *Although it was December, no snow fell and the temperature rose to 20 degrees.*

Here, the primary communicative intention is merely in informing about the events and emphasizing the unusualness of their correlation: The minor act (the "background") is the one that increases the relevance of stating the main act. Optionally, if the reason for the unexpected event is known, it can also be verbalized, as in example (10) above.

# 3 What exactly is a concession?

Given the variety of concession situations and the broad range of possible verbalizations, we need to sort out the different *levels* of description, keeping an eye on the language generation task, to which we will turn in section 5. We first characterize concessions on the *knowledge level* and in the next section will examine their realizations in language.

In spite of the diversity of concessions we found so far, we can extract the following underlying general situation, relating propositions and implications:

> *On the one hand, A holds, implying the expectation of C. On the other hand, B holds, which implies Not-C, contrary to the expectation induced by A.*

More formally, we have a defeasible implication, and a rule induced by the context

```
A  -> C
B  -> Not-C
```

|    | A (although) | B (because) | Not-C (nevertheless) | A -> C | B -> Not-C |
|----|--------------|-------------|----------------------|--------|------------|
| 1  | make-effort(he) | ? | not-president(he) | make-effort(x) → president(X) | ? |
| 2  | old(he) | ? | working(he) | old(X) → not-working(X) | ? |
| 3  | rain | ? | go-for-walk | rain → not-go-for-walk(X) | ? |
| 4  | bad-weather | ? | go-for-walk | bad-weather → not-go-for-walk(X) | ? |
| 5  | pressure | ? | not-order-return | pressure → order-return | ? |
| 6  | brief(papers) | ? | not-bad(papers) | brief(X) → bad(X) | ? |
| 7  | toxic(material, animals) | ? | not-toxic(material, humans) | toxic(X,animals) → toxic(X,humans) | ? |
| 8  | near(i,they) | speak-low (they) | not-hear(i,they) | near(X,Y) → hear(X,Y) | speak-low(X) → not-hear(X,Y) |
| 9  | not-own(he,car) | own(he,bike) | *mobile(X)* | *not-own(X,car) → not-mobile(X)* | *own(X,bike) → mobile(X)* |
| 10 | walk(i,dark) | support(lord,i) | no-fear(i) | walk(X,dark) → fear(X) | support(lord,X) → no-fear(X) |

Figure 2: Analysis of the CONCESSION examples in section 1

and the knowledge that both A and B hold simultaneously. In addition, we decide that the second rule is in the current situation stronger than the first one, hence Not-C follows.[4] C forms the connection between A and B, the point (ore 'value', see Lang 1989) that both A and B relate to, hence allowing for the combination of the two propositions. Typically, the first (default) implication encodes general world knowledge, either a rule of cause and effect, or a customary expectation.

Now, the second rule and the proposition B can be unverbalized in the argument, in which case we are left with stating the violation of the first rule: *Despite the bad weather we had a good time.* There surely were reasons for the good time (B), but they are either unknown or irrelevant and therefore not being communicated. Also, the conclusion C does not have to be explicit in the discourse: *Granted, he failed the exam; but he's a real good speaker.* Whatever the conclusion (e.g., *Let's hire him*), it is not explicitly communicated here, for instance because it is already mutually known.

Exemplary for this type is the case where the contrastive statement denotes a *substitution* of some kind.[5] An, apparently negative, statement is made, and another one serves as — possibly partial — compensation. This case is exemplified with example (9).[6] Examples like these are well-formed only in case the two propositions

---

[4] To be more illustrative, we can call A the "although"-part, B the "because"-part, and C/Not-C the "nevertheless"-part.

[5] See (Stede 1994) for a contrastive analysis of English and German SUBSTITUTION markers.

[6] English gloss: *He doesn't have a car, but he has a bike.*

are connected somehow, or share a "common ground". Compare:

(18) ?*While the moose got lost in Toronto, maple syrup is sweet.*

The existence of the common ground, the `C` part, provides the link to CONCESSION (see figure 2, (9)): The first clause (*He doesn't have a car*) triggers a possible consequence (e.g., *He is not mobile*), which is refuted (at least partially) by the second clause (*He has a bike*). In this sense, the first clause is conceded, but the second one overwrites or at least adjusts the resulting implication.

The `A,B,C`-scheme accounts for the concessions we examined in our corpus studies, despite the variety of definitions that were supposed to underly them. In figure 2, we apply the scheme to the 10 examples given in the beginning. The parts in italics in example (9) are not explicit in the sentence but the inferred common ground. The table shows that only in two cases all three parts (`A, B, C`) are verbalized ((9) and (10)). The remaining parts have to be inferred ¿from the context. For example, (1) might be followed by the sentence *Und zwar, weil er nicht clever genug ist — That's because he isn't clever enough*, giving `B = Not-clever(he)` and `(B -> Not-C) = (Not-clever(x) -> Not-president(x))`.

To summmarize, we have two groups of cases: those verbalizing `A` and `C`, the *violated-implication-concessions*, and those verbalizing `A` and B, the *substitution-concessions*. Both groups cross-classify with the pragmatic CONCESSION classes in the last section; henceforth, we will focus our attention on the first group.

## 4 How do we concede?

We now turn to the opposite end of the problem and look at linguistic realizations of concessions. As opposed to relationships like "background" or "elaboration", concessions are in the vast majority of cases explicitly signalled with cue words, or discourse markers, as we shall call them. We first describe the range of concessive markers and characterize the contexts they are most likely to be chosen in. Then, we turn to a level above and collect observations on the ordering of the CONCESSION parts, and on their syntactic structure.

### 4.1 Lexical realization of CONCESSION

With the help of grammars (Helbig and Buscha 1991, Quirk et al. 1972) and dictionaries (Cobuild 1987, Duden 1989), we gathered a list of German and English markers that can signal CONCESSION and then collected usages from our corpora. Especially German offers a very wide range of particles to mark the CONCESSION relation (see Lenke et al. 1995). Yet, *obwohl* seems to serve as the general-purpose CONCESSION marker, which can often be substituted for others, given that the necessary syntatic changes are performed. Correspondingly, the English *although* is of similar generality.

In order to describe the range of possible discourse markers and their context of occurence as observed in the corpus, we have to impose some kind of classification upon them. Different classifications for concessive markers have been proposed (see e.g. Martin 1992, Halliday 1985, Quirk et al. 1972, Helbig and Buscha 1991), all coming up with different groups of markers. The divergence of the classifications has to do with the point of departure for the analysis, for instance, the grammatical class, stylistic features, pragmatic function in discourse, and the like. For presentational reasons as well as for generation purposes, we will base our description on a classification of concessive markers observed in our corpus according to their function in discourse (cf. Quirk et al. 1972, Halliday 1985).

Classifying the discourse markers according to the kind of bond they create between constituents leads to the following three groups.

- Markers that create a **cohesive** bond by relating a clause to the preceding text. These are above all conjunctive adjuncts[7], most frequently realized by means of an adverb or a prepositional phrase. Coordinating conjunctions fall into this group, too, when occuring in sentence-initial position.

- Markers that form **paratactic** clause complexes, thus creating an **interdependency** relation. These are coordinating conjunctions which form intra-clausal relations. Conjunctive adjuncts can occur within a clause complex, too, but as opposed to conjunctions, they do not create any kind of dependency relation between the clauses involved.

- Markers that form **hypotactic** clause complexes, thus creating a **dependency** relation. There are two options for realization: Subordinating conjunctions, which relate two clauses, and prepositions, which differ from all the other markers in that the relation is not realized between processes but within a process. Hence, it often involves a nominalization of one of the processes, which is given as a circumstance (Martin 1992).

In the following, we will provide a classification of English and German concessive markers encountered in our corpora along these lines. Within the groups, the conditions of usage for the individual markers will only be sketched briefly, for a more detailed account see Lenke et al. (1995).

**Conjunctive adjuncts** encountered in the corpus are *nevertheless, nonetheless, however, still, admittedly*. They are all very similar in meaning, with *however* being the most general one. In the unmarked case, the concessive relation holds between two adjacent clauses; only *still* can be used to connect non-adjacent clauses (Halliday 1985). German offers the following range: *trotzdem* [102][8], *nichtsdestoweniger* [1], *nichtsdestotrotz* [0], *gleichwohl* [21], *dennoch* [94]. See examples (1), (6) and (8) for illustration.

---

[7]The terminology with respect to this syntactic function is slightly confusing; other terms encountered in the literature are: *conjunct* (Quirk et al. 1972), *conjunctive* (Martin 1992), *sentence adverb* (Cobuild 1987), or *Konjunktionaladverb* (Helbig and Buscha 1991).

In both languages, there is a range of adjuncts that is central to the argumentative use. These are the German *wohl, zugegebenermaßen, zugestandenermaßen, freilich, schon, allerdings* und *durchaus*, and the English adjuncts *anyhow, anyway* and the more formal prepositional phrases *in spite of it all, despite all this*. The former are mostly used in dialogues, introducing counterarguments phrased by the speaker as a concession.

As for the positioning of the concessive adjunct in the clause, we observed that adjuncts tend to occur at points in the clause that are significant for the textual organization, which means at some boundary between functional constituents of the clause: *Theme – Rheme; Mood – Residue; clause-initial; clause-final* (Halliday 1985). In German, the positioning of the adjunct is less constrained. However, in both languages, we observed a tendency for the conjunct to be thematic. Conjunctive adjuncts are anaphoric in nature, hence, the clause containing the conjunctive adjunct is positioned at the end of the CONCESSION argument:

(19) *Thomas Hardy spent all but a few years in his native Dorset. His mind, however, went as deep as the Grand Canyon and discovered rockbed truth about men and women.* (COBUILD)

**Coordinating conjunctions** marking a CONCESSION are contrastive in their primary meaning (*but, yet; aber, doch*), but in some contexts they can be understood as signalling a CONCESSION:

(20) *John is poor. But he is happy.* (Quirk et al. 1972)

*But* and *aber* denote a contrast, which can have a concessive meaning aspect to it when the contrast arises from the unexpectedness of what is being said in the second clause. The unexpectedness depends on our presuppositions. Similarly, *yet* and *doch* tend to introduce a comment or remark that is surprising. Obviously, every CONCESSION implies a CONTRAST, and the marker choice can place emphasis on either aspect of the relationship to be conveyed, i.e., be more or less specific.[9] The CONTRAST-markers *but* and *aber* are also used to signal SUBSTITUTION where in English the degree of elision in the clause plays a role (Stede 1994).

**Subordinating conjunctions** for concessions are chiefly *although* and its more colloquial variant *though*, and the German equivalent *obwohl* [180]. Other frequent English alternatives are *even if, even though* and *while*. *Even though* seems to be an intensified *although* (emphasizing the contrast more, compare example (8): *Even though the material ...*), very often substitutable by the latter, whereas *even if* has a conditional meaning aspect to it.

---

[8]The numbers following German lexemes (also on the following pages) give the frequency of occurrence of the respective marker in the LIMAS corpus, thus conveying the range of applicability of a marker.

[9]Knott and Dale (1995) propose a taxonomy that reflects subsumption relationships between discourse relations.

Regarding German, we find in addition to *obwohl* (see example (2)) the following conjunctions: *obgleich* [26], *obzwar* [0], *obschon* [8], *wenngleich* [21], *wiewohl* [5], *gleichwie* [1], *wenn auch* [108]. *Obgleich, obzwar,* and *obschon* are variants of *obwohl*, differing in formality and archaicness.

**Prepositions** that signal a CONCESSION relation within a clause are *despite, in spite of* and *notwithstanding*, with the first two playing the role of the German *trotz* (see examples (4) and (5)). *In spite of* and *trotz* serve as the general-purpose prepositions in concessions, similar to the subordinating conjunctions *although* and *obwohl*. *Despite* is the more formal variant, whereas *notwithstanding* is even more formal and legalistic in style (Quirk et al. 1972). In German, formality can be signalled by the use of *ungeachtet*.

**Split particles** signalling CONCESSION can create inter- and intra-clausal relations. In the English data, we observed *on the one hand – on the other hand*, which is another way of expressing CONTRAST, with the second element being more prominent.

German marker pairs like *zwar+aber* [207], *zwar+doch* [42], *zwar+jedoch* [55], *zwar+ aber+doch* [19], *zwar+other markers* [49] are used for substitution-concessions, predominantly with argumentative goals. There is no literal English equivalent.

(21) *Hans hat zwar viel Schokolade gegessen, aber keine Kekse.*

## 4.2 Syntactic realization of CONCESSION

We now move beyond the word level and turn to observations on the *ordering* of the concessive arguments as well as the *syntactic constructions* by means of which the entire CONCESSION is realized. In our corpus studies, clear tendencies could be observed for every concession type introduced in section 2 regarding the order of clauses in discourse and the syntactic structure realizing them.

**Concede-I** The anticipated counterargument presenting the *given* information is stated first, thereby enabling the rhetorical effect.[10] In this case, hypotactic contructions and consequently subordinating conjunctions will be preferred as they indicate the statement as *given* information. However, if the counterargument is complex, it forms a complete sentence, and the own argument follows, linked by a cohesive element (coordinating conjunction or conjunctive adjunct, cf. section 4.1).

---

[10]This is in line with Halliday's (1985) observation that the *given* information is usually presented before the *new* information.

If the CONCESSION relation is to be especially emphazised, which is typical for the argumentative use, a conjunctive adjunct, like for instance *nevertheless*, can be added to the second clause ( *Although the weather was bad, we nevertheless had a good time.*) Also, the "although" (`A`) clause can be marked with an element like *you are right*, establishing a personal link to the hearer. In the argumentative use, it is central to express one's attitude towards the denied argument. Hence, we do not encounter a hypotactic construction containing a prepositional phrase, since this does not allow for the realization of, e.g., mood and modality. If *evidence* for the own argument is given, it usually follows the argument, forming a clause-complex.

**Concede-II** This class, involving a kind of afterthought, gives the denied implicature after the main act, which is stated in an independent clause. Only then the necessity of denying the possible implicature arises, which yields another independent clause. We observed two possible realizations of the concessive relation: either a conjunctive adjunct is used to create a cohesive bond between two clause complexes, or the clauses are in a paratactic relationship, linked by means of a coordinating concessive conjunction. Recall example (15).

**Concede-III** Here, ordering is less rigid, but we most often find the minor act (the antecedent of the violated default rule (`A -> C`), which increases the relevance of communicating the main act, mentioned first. For heavy emphasis, this order can be reversed, as in this variant of (17):

(22)

(a) *Es fiel kein Schnee und die Temperatur stieg auf 20 Grad, und das, obwohl es Dezember war!*

(b) *No snow fell and the temperature rose to 20 degrees. And that although it was December!*

Again, as with Concede-I, the syntactic realization depends on the complexity of this part. If it is complex, the `A` forms an independent sentence linked to the main act by means of a conjunctive adjunct. As opposed to Concede-I, we find a number of occurrences of prepositional phrases as realization of the `A` ("although") clause. Here, the constraints that disfavour the PP in the case of Concede-I do not apply.

## 4.3 Concession types and surface realization

Our findings in section 4.1 and 4.2 suggest that a specific type of CONCESSION (see section 2) is not signalled by the choice of a particular discourse marker, but rather by the distinct ordering of the concession parts and the syntactic constructions employed to realize them. Hence, there is no direct mapping from the concession type to a single discourse marker. This is due to the fact that concessive discourse markers communicate a more general contrast and failed expectation, which is to various degrees common to all classes of CONCESSION we identified. However, decisions concerning the order of the arguments and the syntax impose a first constraint on the set of markers available to verbalize the CONCESSION. In other words, the corpus studies make us assume a correlation between the concession types and the classification of concessive markers according to their function in discourse, as discussed in section 4.1.

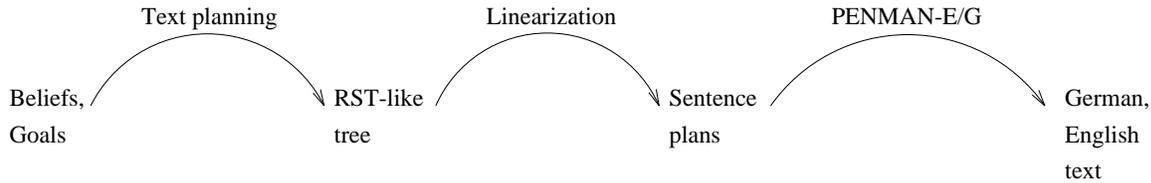

Figure 3: Overall generation architecture

Within these sets, which correspond to the groups defined in section 4.1, the markers are fairly interchangeable, as contrastive studies we performed and the observations from the corpus suggest (see Lenke et al. 1995). Still, we have identified differences between the markers within the groups with respect to intensification, register and stylistic dimensions like formality and archaicness. Furthermore, they can suggest additional meaning aspects, as for instance conditionality (*even if*). In short, the pragmatic goal underlying a concessive statement only determines a group of markers that is defined by a common function in discourse, whereas the final choice of a specific marker depends on factors other than the motivation underlying the CONCESSION.

# 5 Towards generating concessions

Finally, we will put our observations made in the last sections together and describe a framework for actually generating text involving concessions in both English and German. The system is under development; implementation of the systemic networks in the front-end-generator is complete, whereas the other modules are under development.

The sequence of steps is illustrated in figure 3. Starting from a representation of beliefs, attitudes and communicative goals, we build a tree structure following the basic principles of RST (Mann and Thompson 1987): Discourse structure can be represented as a connected tree; most relations assign different status to their relata (nucleus vs. satellite). We do not subscribe to their exact set of relation definitions, though (see below), and we see relations as not holding between clauses as the minimal unit, but rather on the level of propositions (cf. Rösner and Stede 1992).

The tree captures the content that will actually be verbalized in both languages; this is still a language-independent level of representation. In the second step, this tree is being linearized, i.e., transformed into a sequence of semantic specifications corresponding to a single sentence each. For the specifications, we use the Sentence Planning Language (SPL) (Kasper 1989); its expressions are given to the PENMAN sentence generator (Penman 1989) and a German variant developed at FAW Ulm (Grote 1993); for both languages, we are implementing enhancements for expressing concessions.

## 5.1 Beliefs and goals

In section 2, we have proposed three basic classes of concessions, which we can now examine in more detail, thereby laying the foundations for a formal representation of beliefs and goals, from which a generator can start its work. As introduced in section 3, we have the basic scheme of rules and propositions underlying concessions; in addition, the attitudes towards the propositions can be relevant, as well as the beliefs on what the hearer already knows (presuppositions). Thus:

- Knowledge and beliefs of the speaker: Propositions `A, B, C`, a default rule encoding general world knowledge `A -> C` and a more specific implication induced by the context `B -> Not-C` (where `B`, `C`, or `B -> Not-C` can be unverbalized, cf. section 3).

- Speaker's picture of hearer's beliefs (presuppositions): which of `A, B, C, A -> C, B -> Not-C` does the speaker think the hearer believes.

- The attitude that the speaker holds and the hearer is assumed to hold towards the propositions. This will influence the degree of emphasis placed on the concession, but we have not further investigated this yet.

The other central element are the *communicative intentions*; here, we start from Hovy's (1988, p.17ff) "pragmatic goals" and take those that are relevant to the field of conceding. These are: from the group "factual knowledge" the goal *increase knowledge* (of the hearer), and ¿from the group "actions" the goals *activate or deactivate a specific goal in the hearer*. Reformulated for our framework, we end up with three goals:

- (INFORM `X`) – Tell hearer that proposition `X` holds

- (CONVINCE `X`) – Dto., but assuming that hearer holds a belief incompatible with `X`

- (ACTIVATE `Y`), where `Y` is an unrealized action – Prompt hearer to perform action `Y`

1. **relation name:** EXT-CONCESSION
2. **constraints on N:** none
3. **constraints on S:** none
4. **constraints on the N + S combination:** the situation presented in S typically implies a situation incompatible with the situation presented in N
5. **the effect:** R recognizes that a typical implication between the situation in S and that in N does not hold in this specific instance
6. **locus of the effect:** N and S

Figure 4: Definition of new EXT-CONCESSION

As for the classification in section 2, in Concede-I the main act is a CONVINCE or ACTIVATE, if the hearer is to be convinced to perform an action. In both cases, a minor act of INFORMing about a plausible reason is typically performed. In Concede-II, the main act is to INFORM about a proposition; the minor act is the ensuing denial of a possible inference the hearer might make otherwise. In Concede-III, the hearer is merely INFORMed about the contrasting propositions (see figure 5).

## 5.2 Constructing an RSTish tree

As mentioned above, we use the apparatus of RST as discourse representation framework, but we need to make an amendment to the relation inventory for our purposes. The RST-CONCESSION (figure 1) is a "presentational" (or, in other terminology, "internal") relation that reflects an argumentative intention. In addition, we need to cover cases like those in our Concede-III, where the hearer is informed about a correlation of facts violating an expectation. This is quite similar to RST-CONTRAST, but this relation is multinuclear. In our cases we have one part of the relation increasing (or even creating) the relevance for the other, hence the two do not have the same status. We suggest a definition for such a relation in figure 4 and for the time being call it EXT-CONCESSION for "external concession".

The combinations of communicative goals and presuppositions can be mapped to RST relations and thereby a discourse representation constructed. To illustrate, we discuss the "Windows" example in its various functions and realizations, as an expansion of the examples given earlier.

(i) Convince by anticipating counter-argument (Concede I): *Although you are correct that Windows is cheap, I nevertheless wouldn't buy it, because it has many bugs.*

(ii) Convince hearer to take action, in spite of counter-argument (Concede-I): *You are*

|  | S thinks H believes | S's main act | S's minor act | RST tree fragment |
|---|---|---|---|---|
| (i) C-I | A, A -> C | CONVINCE (NOT-C) | INFORM (B) | (CONCESSION (EVIDENCE NOT-C B) A) |
| (ii) C-I | A, A -> C | ACTIVATE (NOT-C) | INFORM (B) | (CONCESSION (MOTIVATION NOT-C B) A) |
| (iii) C-II | A -> C | INFORM (A) | INFORM (B, B->NOT-C) | (CONCESSION A (EVIDENCE NOT-C B)) |
| (iv) C-III | A -> C | INFORM (NOT-C) | INFORM (A) | (EXT-CONCESSION NOT-C A) |
| (v) C-III | A -> C | INFORM (NOT-C) | INFORM (A) INFORM (B) | (EXT-CONCESSION) (CAUSE NOT-C B) A) |

Figure 5: Representations of examples (i)-(v)

right that Windows is cheap, but you really shouldn't buy it, because it has many bugs!

(iii) Prevent false implicature (Concede-II): *Windows is cheap. That doesn't mean I bought it, though, because it has many bugs.*

(iv) Inform about surprising correlation (Concede-III): *Even though Windows is cheap, I would never buy it!* (supposing I'm known as a bargain-hunter)

(v) Inform about correlation and give reason (Concede-III): *Even though Windows is cheap, I would never buy it, because it has many bugs.*

In terms of our scheme of propositions and rules, the situation underlying all these utterances is (with You replacing I in (ii)):

```
A: windows is cheap
B: windows has many bugs
C: I/You buy windows
A -> C: windows is cheap -> I/You buy windows
B -> NOT-C: windows has many bugs -> I/You don't buy windows
```

The cases differ in terms of presuppositions and the speaker's goals, and ¿from them we can determine the corresponding fragments of RSTish trees. The table in figure 5 gives this information, with the RST notation being (Relation Nucleus Satellite). The difference between nucleus and satellite mirrors the distinction between main and minor act: note that the order in (iii) is opposite to the others.

In (i) and (ii) the hearer is to be convinced of NOT-C, with the difference being that in (ii) this is a potential action of the hearer, and in (i) a proposition that the speaker wants the hearer to believe. Therefore, the relation connecting the minor act is RST-EVIDENCE in (i) and RST-MOTIVATION in (ii).

In (iii), we have main act and minor act reversed. They can, in general, map to either EXT-CONCESSION or RST-CONCESSION, depending on the kind of implicature denied. The verbalization will in any case be different from that of (i) and (ii) on the grounds that the main act is merely (INFORM A).

In (iv) and (v), the INFORM-acts together with the presupposition (which indicates the conflict between the statements) give rise to an EXT-CONCESSION between NOT-C and A, and in (v) the additional verbalization of B leads to a CAUSE relation.

As support for introducing the new relation, note that the relations that get combined with RST-CONCESSION (EVIDENCE and MOTIVATION) are also presentational/internal relations, whereas CAUSE is a subject-matter/external relation, like EXT-CONCESSION.

## 5.3 Linearizing the tree

Mapping an RSTish tree to a sequence of sentence plans has been a focus of interest in our earlier work on tree linearization within the TECHDOC generator (Rösner and Stede 1992), and is now being extended towards making better-informed choices on signalling discourse relations.

The central decisions to be made on this step are determining sentence boundaries and deciding on the order in which to present the information (as an RST tree does not include any information on surface ordering). In practice, obviously, these decisions depend also on the surroundings in an actual tree; here we discuss only the simplified cases dealing with tree fragments involving concessions. Going back to the observations on typical orderings for the three concession classes, as stated in section 4.2, and to the groups of markers given in 4.1, we can give the following provisional rules, similar in nature to those proposed by Scott and de Souza [1990] for other relations.

Note that the RSTish tree does not hold *all* information necessary for the further decisions; we need to go back to the belief/goal representation for instance to identify how nucleus and satellite map to A,B,C, in order to find whether a violated-expectation-concession or a substitution-concession is at hand.[11]

**Concede-I** (examples (i), (ii)): Due to the intended rhetorical effect, we assume a fixed order for the information. The counterargument, i.e. the satellite of the RST-CONCESSION, is stated first. If it is complex (has more than one proposition), it forms a separate SPL, yielding a sentence. The second sentence with the own argument contains a conjunctive adjunct, which is specified with the **:conjunctive** slot in the

---
[11]This is somewhat related to the ongoing discussion whether "one tree is enough" for discourse representation (e.g., Moore and Pollack 1992), a question that we leave aside at this point.

```
(rst / rst-concession                (ext / external-concession
   :domain (r2 / rst-evidence            :domain (rst / rst-cause
             :domain (NEVERTHELESS                 :domain (NEVERTHELESS)
               :conjunctive concessive)            :range (BECAUSE))
             :range (BECAUSE))           :range (ALTHOUGH))
   :range (ALTHOUGH))
```

Figure 6: Partial SPL-expressions for example (i) and (v).

SPL, filled with **concessive**.[12] This will prompt the grammar to choose and realize the adjunct.

If the satellite is a single proposition, a hypotactic construction with subordinating conjunction is chosen, and the subordinate clause will be placed in clause-initial position by the grammar. This is the default order that we specified in PENMAN for the RST-CONCESSION relation. If the presence of the relation is to be specially emphasized, which is typical for the argumentative use in this group, a conjunctive adjunct like *nevertheless* can be added to the second clause.

If the nucleus of RST-CONCESSION is an EVIDENCE or MOTIVATION relation, then its satellite (which corresponds to the B element) is stated last; it typically forms a clause complex with the nucleus of that relation (the own argument), linked with a causal connective.

**Concede-II** (example (iii)): The nucleus of the RST-CONCESSION, corresponding to the main act, is stated first. Then, in a separate SPL, the unwanted implicature is denied, and cohesion is created by means of the **:conjunctive** keyword. If the satellite is simple, though, it can be combined with the nucleus in a paratactic clause; therefore, we build a single SPL.

**Concede-III** (examples (iv),(v)): With this class, the order of nucleus and satellite is not fixed, as it depends on the overall theme development of the text (which we leave aside here). Typically, *given* information is clause-initial, and *new* information clause-final. So, if necessary, the default order assumed for EXT-CONCESSION, which is satellite–nucleus, may need to be overwritten; to do this we use the **:theme** keyword in the SPL. This can also be done if special emphasis is to be placed on the surprising correlation. The satellite moves to second position and is subordinated with *und das* in German, or *and that* in English (see example (22)).

Distributing the content over one or two SPLs works as for Concede-I above, except for hypotactic realizations, where we can have either a subordinating clause or a prepositional phrase. If the reason (B) is given, i.e. there is a CAUSE relation, then its satellite needs to be linked like the EVIDENCE and MOTIVATION satellites above.

---

[12]The original PENMAN expects the lexicalized conjunctive here; we have added the facility to specify just a group, from which the grammar can choose later.

Figure 6 gives the results of the linearization step for examples (i) and (v).[13] In effect, we see that the tree linearization step constrains the syntactic category of possible markers, i.e., it selects one or more of the groups presented in section 4.1. The next step, surface realization, can make the final marker choice from that group.

## 5.4 Surface realization

The sequence of SPLs is now one by one passed to the English PENMAN and the German variant developed at FAW Ulm. The original NIGEL grammar would always produce a hypotactic clause complex with the subordinating conjunction *although* in response to an RST-CONCESSION term in the SPL-expression. Correspondingly, the German grammar produced the general-purpose concessive conjunction *obwohl*. As this realization falls far short of capturing all the differences between types of CONCESSION, we have expanded the systemic grammars of the English and the German version by adding more delicate networks to enable the generation of a wider range of concessive markers. They will now realize both RST-CONCESSION and EXT-CONCESSION.

A systemic account of English concessive markers is provided by Martin (1992, p. 200 ff) for what he terms *Internal consequential relations*, which roughly corresponds to the conjunctive adjuncts described in section 4.1, and for *External hypotactical consequential relations*, which is a subgroup of our subordinating conjunctions (see 4.1), though missing the more delicate distinctions. Up to now, there has neither been a systemic account of the other groups distinguished in section 4.1 nor for German concessive markers in general.

In the following, we look at the subordinating conjunctions only, which yield the most interesting networks. We take the feature [concession-dependent] in NIGEL's system CIRCUMSTANTIAL-DEPENDENT-TYPE as the input feature to our more delicate systems. The system networks for German and English hypotactical concessive relations are presented in figures 7 and 8 (simplified for presentational purposes).

The first choice concerns the substitution/violated-implication concession types explained at the end of section 3. This distinction, however, applies only to German; English does not offer any subordinating conjunctions to signal substitution. The THEMATIZATION system makes its decision based on the nucleus/satellite order specified in the SPL.

The CONDITIONALITY system is the same for both languages, and more delicate distinctions are made in the non-conditional case. In English, we now have the option to intensify the concessive relation with *even though*; otherwise, the FORMALITY system makes decisions with respect to formality.

---

[13]The expressions in capital letters (e.g. ALTHOUGH) are placeholder for the respective propositions to be filled in.

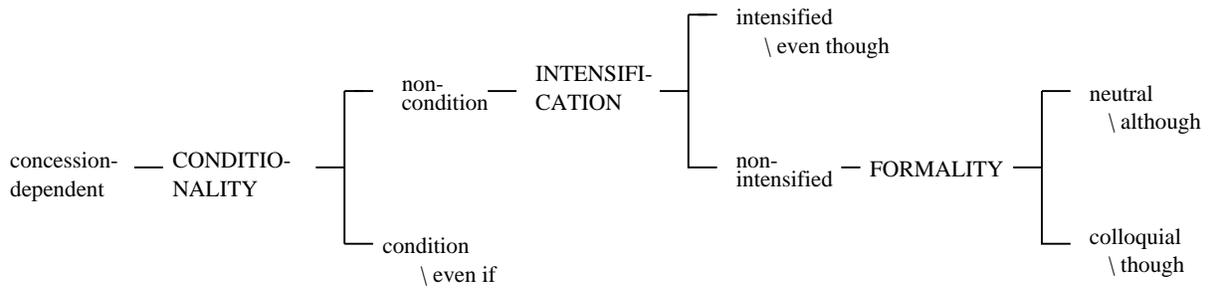

Figure 7: Hypotactical concessive relations in English

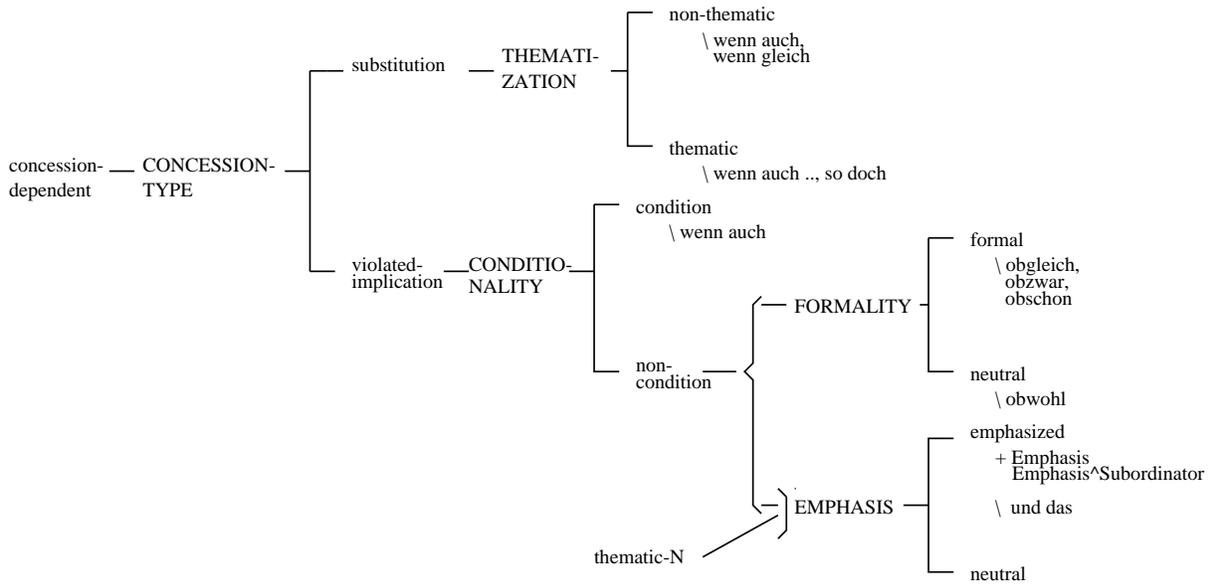

Figure 8: Hypotactical concessive relations in German

In German, FORMALITY and EMPHASIS are traversed in parallel, provided that the [thematic-N] feature has resulted from a thematization of the nucleus in the SPL (see Concede-III in the last subsection). Additional emphasis can then be placed on the satellite by inserting the additional *und das* (example (31)). A choice in the FORMALITY system will be made in any case, and due to the simultaneous systems, we can generate a combined marker like *und das obwohl*. Some features in the German network show more than one realization (e.g., [formal]), as motivations for making these selections could not be drawn from the corpus study.

# 6  Summary and conclusions

The need for linguistic analysis of the discourse markers suitable for use in a NL generation framework was the starting point of our work on making concessions, which ought to be complemented by studying other fields of discourse relations and their markers in similar ways — only then can we get a clearer picture of what exactly the objects are that discourse relations relate. Then we will be able to furnish language generators with knowledge to decide if and how the presence of a relation should be signalled when producing text. Such decisions interact with basically all other choices in text generation, all the way down to lexical selection, where, for instance, choosing a verb determines the (non-) possibility of a nominalization, which in turn constrains the availability of a discourse marker like the German *trotz*. However, in this paper, we have for a start taken one particular field of discourse relation and discussed the work it takes to adequately signal the presence of such a relation, all along from representations of beliefs and goals to surface sentences in both English and German. Faced with the broad range of phenomena that different viewpoints subsume under the heading "concession", we first identified three major classes of *reasons* for conceding. Connecting these with our observations on marker usages, we were lead to propose a classification of the knowledge schema underlying the production of concessions in discourse, and of some parameters that are responsible for their particular verbalizations in different situations. For generation, the crucial task is to classify these parameters in such a way that one can come up with a *serialization* of all the decisions involved. Here, we clearly do not have all the answers yet. It is not at all clear in which order the various parameters influencing this aspect of text generation should be taken into account because they are all interrelated. Nonetheless, a generator needs to impose *some* order at any rate, and we proposed a generation model, with fairly conventional steps and representation levels, that incorporates the task of signalling relations (so far, CONCESSION only).

# Acknowledgements

Examples marked with "COBUILD" are drawn from the Cobuild corpus held at Birmingham University. Thanks to Oliver Jakobs for supplying us with masses of English concordances from that corpus. For helpful comments on earlier versions of this paper, we thank Daniel Marcu, and an anonymous reviewer. The second author's work is funded by the state Nordrhein-Westfalen in the project "Interactions of syntax and semantics". The third author gratefully acknowledges financial support from the Ph.D. scholarship program at FAW Ulm.

# References

[Cobuild 1987] J. Sinclair (ed.). *Collins Cobuild English Language Dictionary*. Collins, London, 1987.

[Duden 1989] *DUDEN: Deutsches Universalwörterbuch*. Dudenverlag, Mannheim 1989.

[Elhadad and McKeown 1990] M. Elhadad, K.R. McKeown. Generating Connectives. In: Proceedings of *COLING-90*, Helsinki, 1990, pp 97-101.

[Glas 1975] R. Glas. Ein Textkorpus für deutsche Gegenwartssprache. In: *Linguistische Berichte* 40, 1975, pp 63-66.

[Grote 1993] B. Grote. Generierung in der systemisch-funktionalen Grammatik: Die Behandlung von Präpositionen. Magisterarbeit, Universität Trier, 1993. (To appear as Technical Report, FAW Ulm)

[Halliday 1985] M.A.K. Halliday. *Introduction to Functional Grammar*. Edward Arnold, London, 1985.

[Halliday and Hasan 1976] M.A.K. Halliday, R. Hasan. *Cohesion in English*. Longman, Burnt Mill, 1976.

[Helbig and Buscha 1991] G. Helbig, J. Buscha. *Deutsche Grammatik*. Langenscheidt / Verlag Enzyklopädie, Berlin, 1991.

[Hovy 1988] E. Hovy. *Generating Natural Language under Pragmatic Constraints*. Lawrence Erlbaum, Hillsdale/NJ, 1988.

[Kasper 1989] R.T. Kasper. A flexible interface for linking applications to PENMAN's sentence generator. In: *Proceedings of the DARPA Workshop on Speech and Natural Language*, 1989.

[Knott and Dale 1995] A. Knott, R. Dale. Using Linguistic Phenomena to Motivate a Set of Coherence Relations. In: *Discourse Processes* 18:1, 35-62, 1994.

[Lang 1989] E. Lang. Syntax und Semantik der Adversativkonnektive. In: B. Kunzmann-Müller, E. Lang (eds.): *Studien zu Funktionswörtern*. Berlin, 1989.

[Lenke et al. 1995] N. Lenke, B. Grote, M. Stede. Concessions and how to make them. Paper presented at the Fourth International Colloquium on Cognitive Science, Donostia-San Sebastian, Spain, 1995.

[Mann and Thompson 1987] W. Mann, S. Thompson. Rhetorical Structure Theory: A Theory of Text Organization. In: L. Polanyi (ed): *The Structure of Discourse*. Ablex, Norwood, 1987.

[Martin 1992] J. Martin. *English Text – System and Structure.* John Benjamins, Philadelphia/Amsterdam, 1972.

[Moore and Pollack 1992] J. Moore, M. Pollack. A problem for RST: the need for multi-level discourse analysis. In: *Computational Linguistics* 18(4)

[Penman 1989] *The Penman Documentation.* Unpublished Documentation of the Penman Sentence Generation System. USC/ISI, Marina del Rey/CA, 1989.

[Quirk et al. 1972] R. Quirk, S. Greenbaum, G. Leech, J. Svartvik. *A Grammar of Contemporary English.* Longman, Burnt Mill, 1972.

[Rösner and Stede 1992] Dietmar Rösner and Manfred Stede. Customizing RST for the automatic production of technical manuals. In R. Dale, E. Hovy, D. Rösner, and O. Stock, editors, *Aspects of Automated Natural Language Generation.* Springer, Berlin/Heidelberg, 1992.

[Scott and de Souza 1990] D. Scott and C. de Souza. Getting the message across in RST-based text generation. In: R. Dale, C. Mellish, M. Zock (eds.): *Current Research in Natural Language Generation.* Academic Press, London, 1990.

[Stede 1994] M. Stede. A contrastive analysis of some contrastive discourse markers. In: J.J. Quantz, B. Schmitz (eds.): Ambiguity and Strategies of Disambiguation (Proceedings of a Workshop held at the Annual Conference of the Deutsche Gesellschaft für Sprachwissenschaft). Technical Report KIT-120, FB Informatik, TU Berlin, 1994.